# ChatGPT as speechwriter for the French presidents

Dominique Labbé, Cyril Labbé, Jacques Savoy

| Science Po | Laboratoire LIG | Computer Science Dept. |
| Univ. Grenoble Alpes | Univ. Grenoble Alpes | University of Neuchatel |
| 1030 av. Centrale | 700 av. Centrale | rue Emile Argand 11 |
| 38400 St Martin-d'Hères | 38000 Grenoble | 2000 Neuchatel, Switzerland |
| Dominique.Labbé@umrpacte.fr | Cyril.Labbe@imag.fr | Jacques.Savoy@unine.ch |

**Abstract**

Generative AI proposes several large language models (LLMs) to automatically generate a message in response to users' requests. Such scientific breakthroughs promote new writing assistants but with some fears. The main focus of this study is to analyze the written style of one LLM called ChatGPT by comparing its generated messages with those of the recent French presidents. To achieve this, we compare end-of-the-year addresses written by Chirac, Sarkozy, Hollande, and Macron with those automatically produced by ChatGPT. We found that ChatGPT tends to overuse nouns, possessive determiners, and numbers. On the other hand, the generated speeches employ less verbs, pronouns, and adverbs and include, in mean, too standardized sentences. Considering some words, one can observe that ChatGPT tends to overuse "to must" (*devoir*), "to continue" or the lemma "we" (*nous*). Moreover, GPT underuses the auxiliary verb "to be" (*être*), or the modal verbs "to will" (*vouloir*) or "to have to" (*falloir*). In addition, when a short text is provided as example to ChatGPT, the machine can generate a short message with a style closed to the original wording. Finally, we reveal that ChatGPT style exposes distinct features compared to real presidential speeches.

## 1 Introduction

Based on large language models (LLMs) (Zhao *et al*., 2023), (Liu *et al*., 2023) trained on huge corpora, the machine could generate short answers to users' requests. The produced messages are clear, coherent, plausible, and without spelling errors. Facing such successes, Bubeck *et al*. (2023) assert that GPT "produces outputs that are essentially indistinguishable from (even better than) what humans could produce". The news however report also other examples of such generated texts that include both incoherence and inaccuracies (called hallucinations). In this context, the current study focuses on the written style adopted by ChatGPT[1] when requesting to produce short political messages.

---

[1] To be precise, GPT is the foundation model employed to generate the next token based on the four previous ones while ChatGPT is the fine-tuned application (based on GPT) to maintain the dialogue with the user.

Currently several LLMs coexist such as Google Bard, Gemini, PaLM 2, Meta Llama 2, but OpenAI's GPT is the most well-known. The initial objective is to produce a chatbot able to maintain a dialogue, for example, to help the user in identifying and resolving a problem. Trained with massive web-text data (e.g., Wikipedia), newspapers[2] and books corpora, the produced messages correspond to short wordings and producing a longer passage required many interactions with the writing assistant. Therefore, the target text genre is a short response within a question/answering system but the produced reply takes account of previous interactions. According to this background, this study will focus on relatively short messages generated by ChatGPT and compared their stylistic features with true ones.

More precisely, the main aim of this study is to empirically test whether ChatGPT can adopt the style of a given author (Savoy, 2020). Moreover, we want to analyze the stylistic differences between the text generated by ChatGPT and the messages written by humans. Do such dissimilarities really occur? If yes, are they observable? In addition, can we characterize some stylistic features used by ChatGPT compared to true authors?

In the remaining part of this article, we will first present some related work while Section 3 exposes the corpus used. Section 4 analyzes some ChatGPT stylistic features by comparing the POS distribution to those occurring in speeches written by French presidents. Section 5 presents analyses based on the vocabularies and the most frequent words occurring in messages written by human and those generated by machine. Section 6 some additional experiments based on the mean sentence length and distribution while Section 7 proposes to verify whether a machine could detect a text generated by ChatGPT. Finally, a conclusion reports the main findings of this study.

## 2   State of the Art

With the recent development of deep learning architecture (Goodfellow *et al.*, 2016), more complex classification and prediction tasks have been solved by different neural networks models. For many NLP applications, the fundamental input representation is a sequence of words (or tokens) to capture the context together with the relative position of each element. As output, a generated sequence is expected (e.g., a translated sentence, a text summary). As a solution to this difficulty, various RNNs (recurrent neural networks) have been suggested (Goodfellow *et al.*, 2016). More recently, transformer models[3] have been proposed to solve effectively such sequence-to-sequence tasks (e.g., BERT, LaMDA, GPT). The underlying idea of such model is to represent the input sequence through a set of encoders with an attention mechanism (Vaswami *et al.* 2017). Based on the internal representation, an output sequence is generated by a set of decoders. To achieve this, both a semantic and a positional representation are taken into account.

---

[2] On Dec. 27th 2023, *New York Times* sues OpenAI and Microsoft for copyright infringement on millions of its articles.

[3] A short tutorial on such models can be found at https://www.datacamp.com/tutorial/how-transformers-work.

To keep things simple (Wolfram, 2023) the foundation model GPT is able, based on a sequence of initial tokens, to produce a ranked list of the next plausible token (e.g., word or punctuation symbol), list elaborated based on the training documents. For example, after sequence "the prime minister of", the model can define a list of the next token as {UK, India, England, Canada, France, Australia, …}.

From this list, and depending on some parameters, the system can then select the most probable one (e.g., "UK" in our case), or based on a uniform distribution, one over the top *k* ranked tokens (e.g., "Canada") or, randomly depending on their respective probabilities of occurrence in the training texts (e.g., "India"). This non-deterministic process[4] guarantees that the same request could create distinct messages. It is important to note that the choice of the next token is reasonable or plausible according to the training sample. This does not mean that the produced sequence does exist in the training documents or that the resulting statement is true. Thus, and common to all LLMs, GPT output may include hallucinations (incorrect information) in its answers. The writing assistant will just provide a reasonable or plausible next token. Moreover, the specification of the sources exploited to produce the text stays unknown to the user.

When working with such writing assistants, can we discriminate between texts written by a machine or a human being? Several studies expose the effectiveness of the learning strategies able to discriminate between responses generated by GPT-3 or written by human beings (Guo *et al.*, 2023). Based on trained black-box classifiers (e.g., RoBERTa), the recognition rate is rather high (around 95% to 98%). Such effectiveness is even obtained when the target language is not English (e.g., French in (Antoun *et al.*, 2023)). Such a high degree could be lower when faced with a new and unknown domain or when substituting tokens by misspelled words (in such cases, the achieved accuracy rate varies from 28% to 60%). In a related study, (Gao *et al.*, 2023) indicate that human beings are less efficient to detect machine-generated passages.

More problematic is the use of such writing assistants to produce scientific papers. Such applications are not new (Labbé & Labbé, 2013) even in generating tortured phrases (Cabanac *et al.*, 2021). According to Gao *et al*. (2023), the scientific abstracts generated by GPT are hard to detect by expert in the field (success rate around 68%, high-impact journals). In this case, GPT abstracts appear vague, superficial, focusing on details and could employ alternative spelling of words. According to (Picazo-Sanchez & Ortiz-Martin, 2024), GPT have been used in around 10% of the 45,000 papers from 3009 journals analyzed by four automatic detectors. Recently, (Soto *et al*. 2024) found that LLM exhibit similar and consistent results.

The main drawback of those LLMs is the need of large number of texts to train them. Of course, their knowledge is time-limited by the period covered by the training set. One can assume that the English language represents the largest part of the training sample, implying a better performance in this language. Moreover, to achieve very good performance, the training sample must be similar to the test one, with similar topics and text genre. When facing with a new domain, text genre or with incorrectly spelled passages (Antoun *et al.*, 2023) (Soto *et al.*, 2024), the performance drops clearly. In addition, such detectors must take account of updated LLMs in response to their known weaknesses. Therefore, there is a clear requirement to acquire a style description or to derive some stylistic features associated to such LLMs and, in this study, for GPT.

---

[4] This random aspect is under the control of the temperature parameter.

## 3  Corpus Overview

To compare the written style of recent French presidents with speeches generated by machine, ChatGPT (chat.openai.com) was asked to generate the end-of-the-year address of four presidents, namely Chirac, Sarkozy, Hollande, and Macron. For each leader and each year, we asked ChatGPT to generate an end-of-the-year address after submitting it the corresponding natural text (NT) as a model.

In our corpus, all NTs and GPTs[5] were corrected and labeled according to the principles specified by Muller (1977). This implies that an orthographic standardization was applied to ensure that any differences observed between NTs and GPTs do not stem from fluctuations in word spelling. Lemmatization adds a label (lemma) to each word in the text, including its dictionary entry (for example, the infinitive of the verb) and its grammatical category. The characteristics of ChatGPT's vocabulary (see Section 4) suggest that ChatGPT does not really know French grammar and all the vocabulary, but that it works with a wide variety of tokens (graphic forms) and seems to have some difficulty with certain homographs.

To have an overview of our corpus, Table 1 provides the president names and $N_{nt}$, the number of words in each presidential message, followed by $N_{gpt}$, the number of words of the corresponding generated text. As we don't know exactly how ChatGPT divides messages into tokens, we have decided to divide them into words. For example, "*aujourd'hui*" (today) or "*parce que*" (because) are single words in French because one can found them as entries in language dictionaries.

As shown in Table 1, on average, the GPTs are almost half as long as the models. Only once (Hollande's message in 2012), the generated text is slightly longer than the natural one. As a result, more than half of these generated texts are less than 1,000 words long. Finally, one notes that the lengths of the texts generated fluctuate considerably, suggesting that, in addition to a certain instability inherent in the method, ChatGPT draws on resources other than the texts submitted by the user.

Two questions arise. First, what are the stylistic and vocabulary idiosyncrasies of the GPTs in relation to the NTs used as models? This question will be examined at three levels, namely the parts-of-speech (POS), vocabulary, and style. Second, can GPTs be detected, or are they close enough to the NTs, provided as models, to escape a detection procedure based on the calculation of intertextual distance and automatic classifications?

---

[5] In this paper, we denote by NTs all natural addresses (or real ones) and by GPTs all messages generated by ChatGPT.

|   | President/Year | President Speeches (NTs) $N_{nt_i}$ | GPTs $N_{gpt_i}$ | $N_{gpt_i}/N_{nt_i}$ |
|---|---|---|---|---|
|   | Chirac |   |   |   |
| 1 | 2002 | 1,026 | 552 | 0.54 |
| 2 | 2003 | 1,179 | 639 | 0.54 |
| 3 | 2004 | 1,304 | 446 | 0.34 |
| 4 | 2005 | 980 | 786 | 0.80 |
| 5 | 2006 | 1,142 | 420 | 0.37 |
|   | Total Chirac | 5,631 | 2,843 | 0.50 |
|   | Sarkozy |   |   |   |
| 6 | 2007 | 1,552 | 822 | 0.53 |
| 7 | 2008 | 1,158 | 770 | 0.66 |
| 8 | 2009 | 1,182 | 530 | 0.45 |
| 9 | 2010 | 1,216 | 693 | 0.57 |
| 10 | 2011 | 1,309 | 1,129 | 0.86 |
|   | Total Sarkozy | 6,417 | 3,944 | 0.61 |
|   | Hollande |   |   |   |
| 11 | 2012 | 1,214 | 1,434 | 1.18 |
| 12 | 2013 | 1,639 | 1,023 | 0.62 |
| 13 | 2014 | 1,547 | 1,213 | 0.78 |
| 14 | 2015 | 1,443 | 509 | 0.35 |
| 15 | 2016 | 1,426 | 1,056 | 0.74 |
|   | Total Hollande | 7,269 | 5,235 | 0,72 |
|   | Macron |   |   |   |
| 16 | 2017 | 2,387 | 850 | 0.36 |
| 17 | 2018 | 2,417 | 705 | 0.29 |
| 18 | 2019 | 2,510 | 787 | 0.31 |
| 19 | 2020 | 2,175 | 552 | 0.25 |
| 20 | 2021 | 2,129 | 1,783 | 0.84 |
|   | Total Macron | 9,818 | 4,677 | 0,49 |
|   | Overall total | 30,935 | 16,699 | 0.54 |

Table 1. Lengths in words of presidents' addresses (NT) and texts generated by machine (GPT)

## 4 Part-Of-Speech

Word labeling by the lemmatizer allows us to observe how ChatGPT employed the main grammatical categories, and whether the characteristics of the generated messages match those of the texts it has been given as models. To provide an answer to this question, we have grouped all the GPTs into a single corpus, which is compared with all the presidential messages merged together. The results of this comparison are displayed in Table 2.

In this table, the first column indicates the part-of-speech subdivided by some morphological information (e.g., present tense for verbs). The second displayed the frequencies in permille in messages written by the presidents ($F_{nt}$) while in the third column ($F_{gpt}$) presents the same information for generated texts. The fourth column shows the percentage of difference compared to NTs.

| POS | $F_{nt}$ | $F_{gpt}$ | $(F_{gpt}-F_{nt})/F_{nt}$ % | S |
|---|---|---|---|---|
| Verbs | 154.1 | 151.9 | -1.5 | 0.009 |
|   Futures | 11.0 | 10.0 | -9 | - |
|   Conditionals | 1.2 | 0.6 | -50 | - |
|   Present tenses | 74.3 | 69.1 | -7 | 0.002 |
|   Imperfect tenses | 2.0 | 0.6 | -71 | 0.000 |
|   Pasts simple | 0.5 | 0.1 | -84 | - |
|   Past participles | 20.5 | 21.5 | +4.8 | 0.977 |
|   Present participles | 4.0 | 6.0 | +50 | 0.999 |
|   Infinitives | 40.7 | 44.1 | + 8 | 0.995 |
| Proper nouns | 16.5 | 14.8 | -10 | 0.015 |
| Common nouns | 193.5 | 207.1 | +7 | 1.000 |
| Adjectives | 56.6 | 64.4 | +13 | 1.000 |
|   Adjectives from past participles | 5.7 | 6.0 | +5 | - |
| Pronouns | 114.7 | 89 | -22 | 0.000 |
|   Personal pronouns | 59.3 | 57.9 | -2.4 | 0.015 |
| Determiners | 197.1 | 210.8 | +7 | 1.000 |
|   Articles | 126.5 | 124.8 | -1 | 0.010 |
|   Numbers | 19.1 | 25.9 | +36 | 1.000 |
|   Possessives | 33.0 | 42.8 | +30 | 1.000 |
|   Demonstratives | 9.7 | 11.1 | +14 | 0.983 |
|   Indefinites | 8.7 | 6.1 | -30 | 0.001 |
| Adverbs | 55.0 | 37.3 | -32 | 0.000 |
| Prepositions | 161.8 | 171.3 | +6 | 1.000 |
| Coordinating conjunctions | 35.3 | 41.6 | +18 | 1.000 |
| Subordinating conjunctions | 14.8 | 11.1 | -25 | 0.000 |

Table 2. Densities of grammatical categories in GPTs compared with presidents' messages (per thousand words)

In the last column of Table 2, an index (denoted S) indicates whether the use in the GPTs differs significantly from that in the NTs. This index varies between zero and one. With a risk of error of less than 5%, we can conclude that overemployment is significant if S > 0.95 and underemployment if S < 0.05. At 1%, these threshold values are 0.99 and 0.01 respectively. The risk of error can therefore be deduced directly from the figures. In the tables below, no figure is given when 0.95 > S > 0.05.

To interpret the values appearing in Table 2, the first row indicates that in the presidential addresses (NTs), the density of verbs is 154.1 per thousand words. In texts generated, it is 151.9, a difference of -1.5%. There is less than a 1% chance of being wrong in asserting that this density is significantly lower in GPTs than in NTs (last column).

Except for verbs (notably futures, past participles, and infinitives), proper nouns (which ChatGPT reuses from the original texts without any addition), and demonstrative determiners (e.g., this), the deviations are significant with a risk of error of less than 1%.

Leading the way in overuse are present participles (an invariable verbal form in French), possessive determiners (my, your, ...), numbers (in fact, the year provided in the prompt, which ChatGPT tends to repeat too often), adjectives, coordinating conjunctions, and common nouns.

On the other hand, the underused categories are all linked to the verb: pronouns, adverbs and subordinating conjunctions. For the verb, rare and complicated tenses and modes are avoided, such as the imperfect tense, or even forgotten altogether, like the past simple tense. The complexity of French verb usage may explain ChatGPT relative parsimony. The same goes for subordinate clauses, which ChatGPT seems to avoid as much as possible. In both cases, this complexity results in multiple inflections (and therefore as many different tokens) and consequently low probabilities of occurrence.

Another example tends to confirm our finding. The main subordinating conjunction is "*que*" (that), an homograph of the pronoun (who, which) that is endowed with a high degree of ubiquity in French sentences. This aspect results again in low probabilities of occurrence for each of the possible combinations.

To achieve a better overview, Table 3 combines the categories into two groups, namely the group of POS linked to the verbs and the group of those linked to the nouns.

| Categories | $F_{nt}$ | $F_{gpt}$ | $(F_{gpt}-F_{nt})/F_{nt}$ % | S |
|---|---|---|---|---|
| Verb group of POS | 338.6 | 289.9 | - 14.4 | 0.000 |
| Noun group of POS | 660.8 | 710.0 | + 7.7 | 1.000 |

Table 3. Comparative densities of noun and verb groups of POS in presidential addresses and texts by ChatGPT (per thousand words)

These significant differences in the two groups suggest ChatGPT's reticence towards verbs (at least in the French version) other than in the simpler forms, namely present tense, past participles or invariable forms (present participles and infinitives). The same comment is valid for other POS linked to the verbs, such as pronouns, adverbs, and complex constructions with subordinates. On the other hand, ChatGPT shows a clear preference for common nouns, adjectives, possessives, and prepositions, probably because their associations with other words are more regular.

# 5 Vocabularies

In this section, the analysis focuses on lemmas (dictionary entries) belonging first to the verb group, then to the noun group. In each case, we compare the most frequently used lemmas in the presidential and generated speeches using the rank and the frequency. Since the two samples are of unequal length, the absolute frequencies are converted into relative ones (expressed per thousand words). The last column of the following tables displayed in percentage the difference in frequencies between the natural texts and the generated ones.

| Presidents (NTs) | | | GPTs | | |
|---|---|---|---|---|---|
| Rank | Lemma | $F_{nt}$ (‰) | Rank | $F_{gpt}$ (‰) | $(F_{gpt}-F_{nt})/F_{nt}$ % |
| 1 | être (to be) | 31.68 | 1 | 22.03 | -30 |
| 2 | avoir (to have) | 17.85 | 2 | 16.94 | -5 |
| 3 | faire (to do) | 6.24 | 4 | 5.53 | -11 |
| 4 | devoir (must or should) | 3.62 | 3 | 7.27 | +101 |
| 5 | vouloir (want) | 3.27 | 24 | 0.95 | -71 |
| 6 | pouvoir (can) | 3.07 | 6 | 2.98 | -3 |
| 7 | vivre (live) | 2.72 | 5 | 3.34 | +23 |
| 8 | savoir (know) | 2.00 | 18 | 1.16 | -42 |
| 9 | prendre (take) | 1.88 | 12 | 1.67 | -11 |
| 9 | permettre (allow) | 1.84 | 28 | 0.87 | -53 |
| 11 | aller (go) | 1.71 | 82 | 0.65 | -62 |
| 12 | dire (say) | 1.29 | 91 | 0.29 | -78 |
| 13 | protéger (protect) | 1.13 | 14 | 1.45 | +28 |
| 14 | continuer (continue) | 1.10 | 8 | 2.69 | +145 |
| 15 | penser (think) | 1.10 | 64 | 0.44 | -60 |
| 16 | agir (act) | 1.07 | 20 | 1.09 | +2 |
| 17 | tenir (hold) | 1.07 | 7 | 2.76 | +158 |
| 18 | ouvrir (open) | 1.03 | 73 | 0.36 | -65 |
| 19 | venir (come) | 1.03 | 18 | 1.16 | +13 |
| 20 | falloir (must) | 1.00 | 136 | 0.07 | -93 |

Table 4. Verbs most frequently used by presidents compared with GPT (frequencies per thousand words)

## 5.1 Verb Group

The verb is the backbone of this group. Table 4 gives the rank and frequency of the twenty verbs most frequently used by presidents. The fourth and fifth columns provide the rank and frequency of these twenty lemmas in the GPTs. The last column shows the difference between the two frequencies. If the imitation provided by ChatGPT is perfect, both the ranks and frequencies should be identical and there should be zero everywhere in this last column.

Looking at the first row of Table 4, among presidents, the most frequently used verb is "to be" with a frequency of 31.68 per thousand words. In texts generated by ChatGPT, it is also the first, but with a lower frequency of 22.03‰, given a difference of -30% compared with presidents as displayed in the last column.

In any natural French text, "*être*" (to be), "*avoir*" (to have) and "*faire*" (to do) are the three most frequently used verbs in that order. In GPTs, "*devoir*" (which has several meanings in French (a legal or moral obligation, or a probable achievement in the future)) takes the place of "to do". Moreover, its frequency doubles in GPTs compared with NTs (while "to do" drops by 11%). In presidential addresses, it is the modality of the probable ("next year should be better").

ChatGPT has a little trouble with simple verbs, but even more with verbal compositions (such as "*pouvoir faire*" (to be able to do)). Three cases are particularly striking. First with "to go" (-62%) which is used in French to indicate the near future, that is the main subject of an end-of-the-year message in expressions such as "the next year is going to be better", "some event is going to happen next year", etc. Second, the verb "*falloir*" (must) has almost completely disappeared (-93%). Third, one can notice the considerable decline of "*dire*" (to say) (-78%). Among presidents, the majority of uses are in expressions such as "*il faut dire*" (one must say), "*ce qui* (*ne*) *veut* (*pas*) *dire*" (that is (not) to say), etc.

In these three cases, the main difficulty stems from the impersonal pronoun of the third person "*il*" (it), which is the obligatory subject of "must" (also frequent in front of "to go") and is often found after a punctuation or a subordinating conjunction "*que*" (that) (another difficulty for ChatGPT).

Clearly, ChatGPT has a poor grasp of this type of complex constructions, no doubt because their great variability results in low probabilities of occurrence for each of these constructions used by the presidents. As a result, the same difficulties are amplified for pronouns.

To analyze this question, Table 5 depicts the most frequently used pronouns. One can see that the pronoun "we" is employed the most by presidents (16.1 per thousand words) and by ChatGPT (24.28‰), an increase of 51% compared to presidents.

- 9 -

| | Presidents (NTs) | | GPTs | | |
|---|---|---|---|---|---|
| Rank | Lemma | F$_{nt}$ (‰) | Rank | F$_{gpt}$ (‰) | (F$_{gpt}$-F$_{nt}$)/F$_{nt}$ % |
| 1 | nous (we, us) | 16.10 | 1 | 24.28 | +51 |
| 2 | je (I, me) | 15.97 | 2 | 14.98 | -6 |
| 3 | qui (who, which) | 13.58 | 3 | 10.47 | -23 |
| 4 | ce (it, this one) | 11.35 | 6 | 4.07 | -64 |
| 5 | il (he, she, it) | 8.54 | 9 | 3.27 | -62 |
| 6 | vous (you) | 7.89 | 4 | 8.72 | +11 |
| 7 | se (himself, herself…) | 5.88 | 5 | 4.87 | -17 |
| 8 | que (that) | 5.33 | 7 | 4.07 | -24 |
| 9 | le (him, her, it) | 4.14 | 11 | 1.38 | -67 |
| 10 | celui (this one) | 3.85 | 8 | 2.69 | -30 |
| 11 | tout (everything) | 2.94 | 10 | 2.47 | -16 |
| 12 | chacun (eachone) | 2.59 | 12 | 1.31 | -49 |
| 13 | ils (they) | 2.46 | 13 | 0.65 | -74 |
| 14 | y (there) | 2.26 | 14 | 0.87 | -62 |
| 15 | en (it, some) | 2.20 | 16 | 0.80 | -64 |
| 16 | cela (this) | 1.13 | 14 | 0.95 | -16 |
| 17 | autre (other) | 0.84 | 18 | 0.36 | -57 |
| 18 | on (one) | 0.84 | 29 | 0.07 | -92 |
| 19 | dont (whose, of which) | 0.78 | 18 | 0.36 | -54 |
| 20 | lequel (who, which) | 0.78 | 17 | 0.65 | -17 |

Table 5. Pronouns most frequently used by presidents compared with GPT (frequencies per thousand words)

Table 5 arises the following general two comments. Firstly, apart from "*on*" (one), which leaves GPTs' list of the first twenty pronouns, the other nineteen are common to both lists. ChatGPT's' technique is therefore quite close to the provided speech. But as mentioned earlier in the analysis of grammatical categories, in the texts generated by ChatGPT, there are 22% fewer pronouns than in the texts it is supposed to emulate. Consequently, if ChatGPT had respected the hierarchy of presidential pronouns, the last column should contain around -22% in each row.

This aspect highlights the overuse of "we" and "you". The study of contexts shows that "we" and "you" are mostly used not as subjects of a verb, but as complements ("I present you my best wishes") or as subjects of a relative proposition. In addition, they have no homographs.

Second, the personal pronouns "he" and "they", of which "*elles*" and "*elle*" (she) are flexions, and the pronoun "*on*" (one) are generally subject of a complicated verb such as "*falloir*" (must), "*aller*" (to go) or "*venir*" (to come). Thus, one can see them at the head of a sentence or after an internal punctuation. This could explain their high level of under-use.

| Presidents (NTs) | | | GPTs | | |
|---|---|---|---|---|---|
| Rank | Lemma | $F_{nt}$ (‰) | Rank | $F_{gpt}$ (‰) | $(F_{gpt}-F_{nt})/F_{nt}$ % |
| 1 | plus (more) | 7.44 | 1 | 6.62 | -11 |
| 2 | ne (don't) | 7.34 | 2 | 3.85 | -48 |
| 3 | pas (not) | 4.20 | 4 | 2.47 | -41 |
| 4 | aussi (also) | 3.49 | 10 | 0.87 | -75 |
| 5 | où (where) | 1.52 | 8 | 1.02 | -33 |
| 6 | encore (still) | 1.10 | 15 | 0.51 | -54 |
| 7 | beaucoup (many) | 1.07 | 8 | 0.22 | -79 |
| 7 | tout (all) | 1.07 | 6 | 1.24 | +16 |
| 9 | alors (then) | 1.03 | 18 | 0.44 | -57 |
| 10 | enfin (finally) | 1.00 | 14 | 0.58 | -42 |
| 10 | là (there) | 1.00 | 28 | 0.22 | -78 |
| 10 | toujours (always) | 1.00 | 11 | 0.80 | -20 |
| 13 | bien (well) | 0.94 | 20 | 0.36 | -62 |
| 13 | mieux (better) | 0.94 | 20 | 0.36 | -62 |
| 15 | davantage (more) | 0.84 | 15 | 0.51 | -39 |
| 15 | jamais (never) | 0.84 | 18 | 0.44 | -48 |
| 17 | même (even) | 0.74 | 37 | 0.07 | -91 |
| 18 | aujourd'hui (today) | 0.71 | 13 | 0.73 | +3 |
| 18 | pourquoi (why) | 0.71 | 28 | 0.22 | -69 |
| 20 | ensembe (together) | 0.68 | 5 | 1.31 | +93 |
| (…) | | | | | |
| 37 | également (also) | 0.29 | 3 | 3.71 | +1 178 |

Table 6. Adverbs most frequently used by presidents compared with ChatGPT
(frequencies per thousand words)

In addition, "*ce*" (that), "*le*" (relative pronoun of the third person), "*tout*" (every), "*en*" (other relative pronoun), "*autre*" (other), are homographs of determiners or prepositions. This means a single token for two lemmas and many possible constructions. Moreover, these determiners and prepositions are much more frequent in French than their pronoun homographs.

Clearly, ChatGPT drastically avoids these difficulties. This is reflected in the use of adverbs, the last grammatical category in the verb phrase. Table 6 presents them according to their occurrence frequencies in presidential speeches. From this, the first 15 are present in both lists but not in the same order. One must also recall that this category is clearly under-use by ChatGPT, on average - 32% compared to the natural texts used for learning (see Table 2). This proportion should be borne in mind when interpreting the last column of Table 6.

In any French text, the negative construction ("*ne*+verb+*pas*" (verb+not)) is extremely frequent. It is almost halved in ChatGPT-generated texts, because it is employed exclusively with a verb whose subject is often a pronoun like "*ce n'est pas*" (it is not). On the other hand, adverbs usually used outside the verbal group (e.g., "today") or in a nominal group (e.g., "all") do not suffer the same erosion or are even overused (such as "all" or "together"). The most striking case is "*également*" (also), which ranks third among ChatGPT's favorite adverbs, whereas it is only 37th among presidents. In fact, all its occurrences are found in a nominal group (generally in front of an adjective). However, as the analysis of grammatical categories has shown, ChatGPT's favors words belonging to the noun group, such as these adverbs (the names of which should better be "*adnouns*").

**5.2 Noun Group**

As in the previous subsection, the analysis focuses on the twenty common nouns most frequently used by presidents, with their rank and frequency in GPTs. This information is displayed in Table 7. From this, one can see that in presidential addresses, the most frequently used noun is "year" with a frequency of 5.04 per thousand words. In GPTs, it is also the first but with a frequency of 8.7‰, i.e. +76% compared to presidents.

To have a better interpretation of the values depicted in Table 7, one must remember that ChatGPT uses too many nouns (+7%, see Table 2) compared to presidents. It is this average level that should be kept in mind when analyzing the last column of the table. Thus, the increase of "country" would not be significant. With this in mind, all the other fluctuations are large.

The table shows that the general framework of presidential messages is indeed in the GPTs, but with a stronger emphasis on generic and recurrent theme. For example, "wishes for the year, the future, the country or the nation" are all over-used but, strangely, the fact that these greetings are presented on the evening (December 31st) was not use by ChatGPT (e.g., "*soir*" (evening) – 79%).

In the bottom of the table, the word "challenge" is the second most common noun used by ChatGPT, whereas it occupies a modest place among presidents (72nd place, with a frequency seven times lower). Close examination of the contexts of use shows that in six cases out of ten the construction is "verb + *les* (or *des*) *défis*" (e.g., take up the challenges) and in two cases out of ten, one can see "*face aux défis*" (facing the challenges). In four cases out of ten, challenge is followed by a relative pronoun (which or what). The typical formula is therefore "*relever les défis qui*" (take up the challenge which). This rather stereotyped use may explain why it is repeated so frequently by ChatGPT, but it also proves that it draws some elements from outside the prompted text. He is well aware that French presidents like to talk about "challenges", but he is going the extra mile!

This is confirmed by the last two lines of Table 7. The presidents have all addressed their "fellow citizens" (*concitoyens*). But in French, nouns have a gender, and usage (at least politically correct usage) would dictate that the feminine form be stated before the masculine one. So GPT corrects the presidents: "*nos concitoyennes et nos concitoyens*" (our fellow citizen women and our fellow citizen men)!

| | Presidents (NTs) | | GPTs | | |
|---|---|---|---|---|---|
| Rank | Lemma | $F_{nt}$ (‰) | Rank | $F_{gpt}$ (‰) | $(F_{gpt}-F_{nt})/F_{nt}$ % |
| 1 | année (year) | 5.04 | 1 | 8.87 | +76 |
| 2 | pays (country) | 3.04 | 3 | 3.27 | +8 |
| 3 | compatriote (compatriot) | 2.42 | 13 | 2.11 | -13 |
| 4 | emploi (job) | 2.17 | 5 | 2.69 | +24 |
| 5 | vie (life) | 2.07 | 17 | 1.74 | -16 |
| 6 | monde (world) | 1.94 | 7 | 2.47 | +27 |
| 7 | travail (work) | 1.88 | 18 | 1.60 | -15 |
| 8 | avenir (future) | 1.68 | 6 | 2.54 | +51 |
| 9 | république (republic) | 1.52 | 4 | 2.91 | +91 |
| 10 | gouvernement (government) | 1.45 | 41 | 0.87 | -40 |
| 11 | soir (evening) | 1.36 | 169 | 0.29 | -79 |
| 12 | nation (nation) | 1.29 | 10 | 2.25 | +74 |
| 13 | crise (crisis) | 1.26 | 16 | 1.89 | +50 |
| 14 | face (face) | 1.26 | 10 | 2.25 | +79 |
| 15 | réforme (reform) | 1.23 | 18 | 1.60 | +30 |
| 16 | voeu (wish) | 1.20 | 13 | 2.11 | +76 |
| 17 | mois (mounth) | 1.16 | 73 | 0.58 | -50 |
| 18 | service (service) | 1.13 | 41 | 0.87 | -23 |
| 19 | besoin (need) | 1.10 | 103 | 0.44 | -60 |
| 20 | confiance (trust) | 1.07 | 49 | 0.80 | -25 |
| … | … | … | … | … | … |
| 72 | défi (chalenge) | 0.52 | 2 | 4.14 | +696 |
| 108 | concitoyen (man fellow citizen) | 0,39 | 21 | 1.31 | +238 |

Table 7. Common nouns most frequently used by presidents compared with GPTs (frequencies per thousand words)

Conversely, under-use of the noun "*soir*" can be explained in part by the difficulties ChatGPT encounters with some homographs such as "*ce*" that could be a pronoun (that) or a demonstrative article (this). Presidents always say "*ce soir*" (this evening), so GPT makes very little use of "*soir*", because after "*ce*", there is a greater likelihood of encountering a verb like "*sera*" (that will be), "*sont*" (those are), and so on. Of course, a grammatical analysis would easily separate the determiner "*ce*" (this) from the homograph pronoun (that), but it seems that ChatGPT struggle with this.

Associated with nouns, one can find the adjectives and Table 8 exposes that among presidential addresses, the most frequently used is "new" with a frequency of 2.46 per thousand words. In GPTs, it is also the first, but with a frequency of 2.76‰, +12% compared to president. One must keep in mind that ChatGPT uses 13% too many adjectives compared with presidents (see Table 2). With this information, one can infer that increases in density of "new" or "public" are not really significant.

When combining this table with the previous one, one can see that ChatGPT overuses the main presidential syntagms: "wishes for the new year", "better economic and social situation".

| Presidents (NTs) | | | GPTs | | |
|---|---|---|---|---|---|
| Rank | Lemma | $F_{nt}$ (‰) | Rang | $F_{gpt}$ (‰) | $(F_{gpt}-F_{nt})/F_{nt}$ % |
| 1 | nouveau (new) | 2.46 | 2 | 2.76 | +12 |
| 2 | cher (dear) | 2.39 | 1 | 3.78 | +58 |
| 3 | grand (large) | 2.07 | 7 | 1.09 | -47 |
| 4 | social (social) | 1.97 | 3 | 1.89 | -4 |
| 5 | fort (strong) | 1.36 | 4 | 1.24 | -9 |
| 6 | dernier (last) | 1.10 | 7 | 1.09 | -1 |
| 7 | européen (European) | 1.03 | 17 | 0.87 | -16 |
| 8 | français (French) | 0.94 | 21 | 0.73 | -22 |
| 9 | national (national) | 0.84 | 11 | 1.02 | +21 |
| 9 | public (public) | 0.84 | 12 | 0.95 | +13 |
| 11 | économie (economic) | 0.81 | 7 | 1.09 | +35 |
| 12 | seul (lonesome, alone) | 0.74 | 24 | 0.65 | -12 |
| 13 | bon (good) | 0.71 | 26 | 0.51 | -28 |
| 14 | prochain (next) | 0.71 | 33 | 0.36 | -49 |
| 15 | essentiel (essential) | 0.58 | 5 | 1.16 | +100 |
| 16 | difficile (difficult) | 0.55 | 12 | 0.95 | +73 |
| 17 | fier (proud) | 0.52 | 27 | 0.44 | -15 |
| 18 | nécessaire (necessary) | 0.48 | 18 | 0.80 | +67 |
| 19 | indispensable (indispensa- | 0.45 | 83 | 0.07 | -84 |
| 20 | meilleur (best) | 0.45 | 5 | 1.16 | +158 |
| (…) | | | | | |
| - | vaillant (valiant) | 0 | 44 | 0.29 | ∞ |

Table 8. Adjectives most frequently used by presidents compared with GPTs
(frequencies per thousand words)

At the bottom of Table 8, an adjective used by ChatGPT but absent from the presidential addresses: "*valiant*". In twelve messages, the presidents refer to "our soldiers", and ChatGPT corrects them to "our valiant soldiers". In its framework of words, the usage is to associate "soldier" with "bravery" but this idea is not explicitly present in presidential discourses.

On the other hand, ChatGPT seems to have difficulty handling adjectives that are also nouns (e.g., French, European, neighbor), as well as adverbs "*grand*", "*bon*" (big, well) or the pronoun "*seul*" (alone).

This problem of homographs (which is massive in French) also affects determiners as displayed in Table 9. This table indicates that in presidential addresses, the determiner most frequently used is the article "the" with a frequency of 110.73 per thousand words. In GPTs, it is also the first with an almost equal frequency (110.07‰). The difference between the two (-1%) is not significant.

| | Presidents (NTs) | | GPTs | | |
|---|---|---|---|---|---|
| Rank | Lemma | F$_{nt}$ (‰) | Rank | F$_{gpt}$ (‰) | (F$_{gpt}$-F$_{nt}$)/F$_{nt}$ % |
| 1 | le (the) | 110.73 | 1 | 110.07 | -1 |
| 2 | notre (our) | 18.3 | 2 | 29.08 | 59 |
| 3 | un (one) | 15.75 | 3 | 14.76 | -6 |
| 4 | ce (this) | 9.73 | 4 | 11.12 | 14 |
| 5 | tout (all) | 5.11 | 8 | 3.56 | -30 |
| 6 | mon (mine) | 5.04 | 6 | 6.69 | 33 |
| 7 | son (his) | 4.46 | 11 | 2.18 | -51 |
| 8 | deux (two) | 4.3 | 5 | 6.98 | 62 |
| 9 | mille (thousand) | 4.04 | 7 | 6.54 | 62 |
| 10 | leur (their) | 3.1 | 9 | 2.47 | -20 |
| 11 | votre (your) | 2.13 | 10 | 2.40 | 13 |
| 12 | premier (firs) | 1.55 | 18 | 0.73 | -53 |
| 13 | vingt (twenty) | 1.23 | 12 | 1.60 | 30 |
| 14 | dix (ten) | 1.13 | 13 | 1.53 | 35 |
| 15 | chaque (every, each) | 1.03 | 22 | 0.51 | -51 |
| 16 | huit (eigth) | 0.71 | 14 | 1.09 | 54 |
| 17 | cinq (five) | 0.61 | 15 | 0.95 | 55 |
| 18 | même (same) | 0.58 | 19 | 0.58 | 0 |
| 19 | neuf (new, nine) | 0.52 | 22 | 0.51 | -2 |
| 20 | trois (three) | 0.52 | 16 | 0.80 | 54 |

Table 9. The twenty most frequent determiners by presidents, compared with GPTs (frequencies per thousand words)

The two corpora share the first four determiners in the same order. Thereafter, the differences are quite significant. Most notable is the overuse of "*notre*" (our), "*mon*" (mine) and, to a lesser extent, "*votre*" (your) in connection with the overuse of the pronouns "*nous*" (we) and "*vous*" (you). Similarly, the under-uses of "*son*" (his) and "*leur*" (their) are linked to the low frequencies of third-person pronouns. This is also the case for indefinite pronouns. GPT has difficulty using determiners in the same category, here: "*tout*" (all), "*chaque*" (every) and further down the list, "*autre*" (other). Finally, in GPTs, there is excessive repetition of the date (two thousand...).

To conclude on vocabulary, ChatGPT has difficulties to reproduce NTs for words that often appear after punctuation, with those that may belong to several grammatical categories (homographs), especially when some uses are associated with the verb and others with the noun, and finally with complex constructions such as the "verb+verb" or subordinate clauses.

## 6 Sentence Analysis

It is well-kwon that the style of an author can be measured by several indicators (Savoy, 2020). In this study, after focusing on the POS distribution and the most frequently used lemmas, the analysis of the sentence length can reveal some pertinent aspects of ChatGPT's style. More precisely, one can ask whether ChatGPT succeed in producing sentences that are formally similar to those of the author it has been given as a model? To answer this question, the set of presidents is compared with the set of texts generated by ChatGPT (see also (Monière *et al.*, 2008)). Our analysis is restricted to the sentence lengths.

### 6.1 Sentence Lengths

In the first analysis, sentences are ordered by increasing lengths, and the numbers of each length are counted in both corpora. This yields to two sets of values, summarized in Table 10. In addition, Figure 1 shows the distribution of those lengths for both corpora.

To obtain an overview, Table 10 shows that in the corpus of presidential addresses, the most frequent length (mode) is 12 words versus 19 in the GPTs i.e., 46% longer in the generated texts. Half of the sentences are less than 20 words long (median) in both corpora. The average sentence length is very close (21 and 21.7 words).

|  | Mode | Median | Mean | Standard dev. | CV% | Medial | $(D_9-D_1)/D_1$ |
|---|---|---|---|---|---|---|---|
| Presidents | 13 | 19.9 | 21.0 | 16.4 | 78.1 | 26.4 | 3.84 |
| GPTs | 19 | 20.1 | 21.7 | 10.9 | 50.3 | 24.2 | 1.86 |

Table 10. Main representative values for sentence lengths in the presidential addresses compared to GPTs

Another pertinent information is the medial value. For the presidential speeches, the medial is 26 words, meaning that half of the presidential message is covered by sentences of 26 words or less. For GPT, the medial is lower with to 24 words i.e., 8% less. This medial length is important because it corresponds to durations. For half of the presidential speeches, viewers are confronted with relatively complex sentences whose length (26 words or more) probably exceeds the comprehension capacity of some listeners. Finally, in the presidential texts, the sentence length at the upper limit of the ninth decile is 3.84 times greater than that at the first decile, i.e., more than twice as long as the same interval in the GPTs.

The inequality and spread of central values of sentence lengths is a characteristic of any natural text. To measure this, two indices have been calculated.

First one can consider the standard deviation around the mean. In the presidential messages, around two-thirds of sentences have lengths within the interval (21 ± 16.4) words, whereas in the GPTs, this interval is only (21.7 ± 10.9), i.e., one-third less than in the texts that served as models for GPTs. As a variant, one can compute the coefficient of variation (CV%) as the ratio between the standard deviation and the mean. In presidents' addresses, this value is 16.4 / 21.0 = 78.1% of the mean vs. 50.3% in the GPTs.

Second, the ratio between sentence lengths at the upper bounds of the first and ninth deciles, i.e., an interval comprising 80% of the text (by cutting both ends of the distribution). In the presidential texts, this interval is {12.0 - 57.9 words}, i.e., a ratio of 1 to 3.84. This can be compared with GPTs with an interval {14.4 - 41.2 words} meaning a ratio of 1 to 1.88 or half that of the presidents.

In other words, the generated sentence lengths are significantly less diverse or spread out than in natural texts. The generator has difficulties deviating from the average and producing very short or very long sentences that are familiar in natural language.

**6.2 Sentence Length Distribution**

As the two corpora are not of the same lengths, the absolute numbers of each sentence length are converted into percentages. For example, in Figure 1, there are no one-word long sentence (exclamatory sentence) in the GPTs, compared with 0.27% in the presidential corpus, and so on.

The slight irregularities in the curve, which can be explained by the relatively small size of the two corpora, are not important. Only count the profiles of the two distributions. For example, the two main modes are clearly visible and offset (13 words in the presidential texts and 19 in the GPTs, see also Table 10).

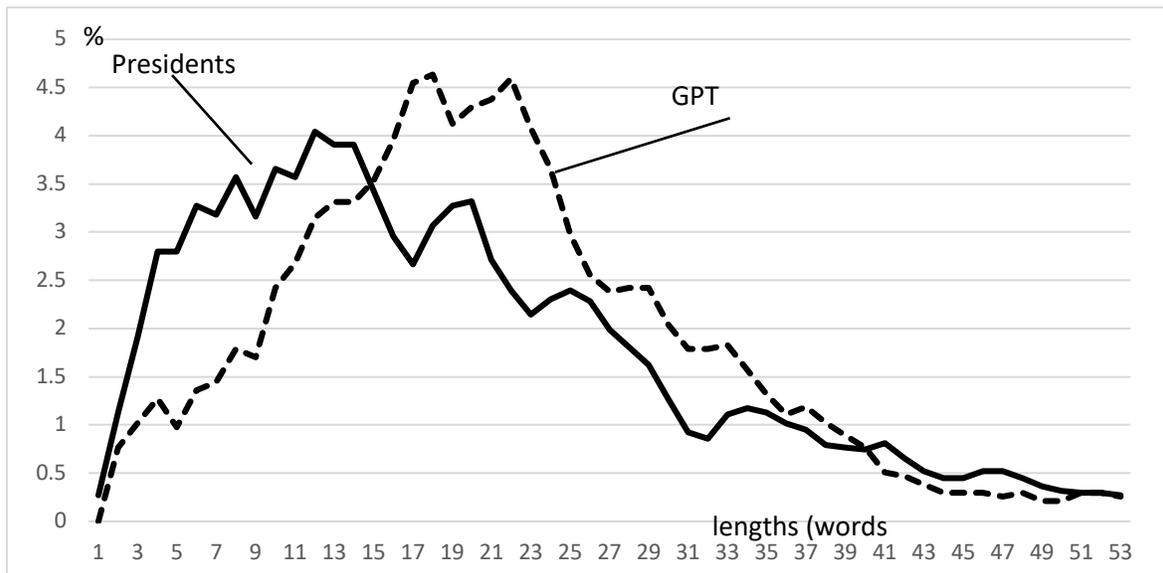

Figure 1.  Comparison of sentence lengths in messages from presidents and those generated by ChatGPT (in % of total sentences in each corpus)

The two distributions are very dissimilar. The presidents (solid curve) show the typical profile seen in any natural text, namely a strong asymmetry on the left and a significant spread of lengths. The dotted curve produced by ChatGPT falls below this solid curve at both ends of the graph, and clearly above it in the middle. Therefore, ChatGPT over-employs "average" sentences (from 15 to 39 words) and avoids "extraordinary" sentences namely fewer very short sentences (lengths less than 15 words) or very long ones (more than 39 words). ChatGPT respects the mean of sentence lengths in the texts it is given as models, but randomly distributes these lengths around the mean, as evidenced by the near-equality of the three central values (mode, median, mean), a characteristic of a Gaussian distribution. On the other hand, the inequality {mode < median < mean} is the main characteristic of sentence lengths in natural texts (either written or oral).

Furthermore, the coefficient of relative variation (CV% in Table 10) indicates that this dispersion around the mean is lower in the corpus of generated texts than in that by the presidents. This finding explains why, on the figure, the mode of sentences produced by ChatGPT is clearly higher than that of natural sentences. Finally, the absence of very long sentences in GPTs may be linked to the low frequency of complex constructions, notably subordinate clauses.

About style, a conclusion is obvious namely that ChatGPT treats punctuation like words (hence the roughly identical central values). Therefore, the generator had identified which words are most likely to be followed by a comma, period, etc.

## 7 Can One Identify a Text Generated by ChatGPT?

To date, the intertextual distance associated with automatic classifications has proved to be an effective tool for automatic recognition of the author of a text. In particular, this method has been used to identify papers produced by previous-generation of paper mills, as well as various types of fraud in scientific publications (Byrne & Labbé, 2016), (Van Noorden, 2014), (Labbé & Labbé, 2012) or well as in authorship attribution (Savoy, 2018).

The calculation compares the vocabulary of two texts (A, B), measuring the absolute difference between the frequencies of each lemma in A and B. The sum of these differences is related to the total length of the two texts. This ratio is the distance between A and B. It varies uniformly between zero (no difference) and one (no lemma in common). For example, a distance of 0.25 means that a quarter of the vocabulary used in A and B is different. Below this threshold, the hypothesis of a single author (for two contemporary texts written in French) can be examined favorably.

In principle, intertextual distance cannot be calculated on texts that are too short, or at least not shorter than 1,000 words. However, ChatGPT first characteristic is to generate short texts (see Table 1). Therefore, to include all messages despite their short lengths, we have merged all generated texts and all natural ones for each president. This results in eight different texts (see Table 1) of which distances had been calculated, enabling to perform the classifications presented in Figure 2 and 3.

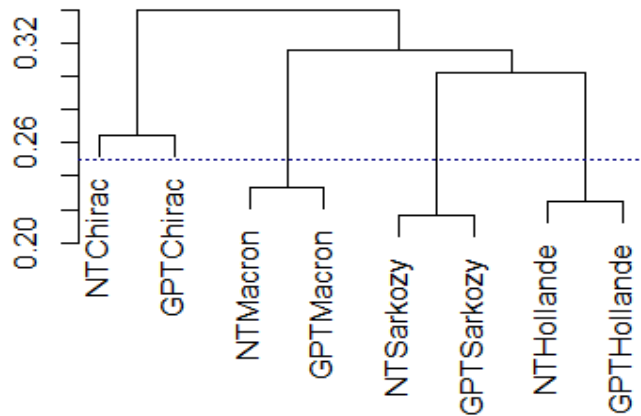

Figure 2. Hierarchical ascending classification by presidents

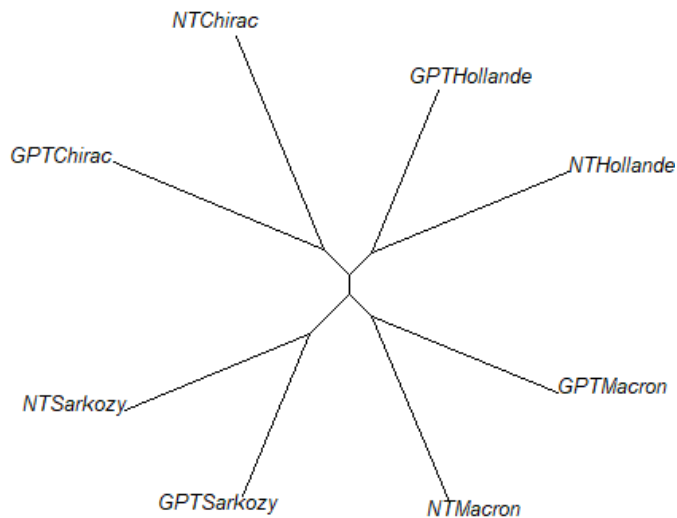

Figure 3. Tree classification by presidents

How much confidence can one place in these classifications? A quality index is used to answer this question (Labbé & Labbé, 2006). For the tree (Figure 3), all the paths on the tree have an index greater than 90%, and the quality of the whole tree is greater than 96%. In other words, there is less than a 5% chance of being wrong in saying that this tree gives the best possible representations of the information contained in the distance matrix.

These two figures provide additional information on the relative proximity of each president. Chirac's messages are quite clearly out of step with those of his three successors. Hollande and Sarkozy are very close; Macron is closer to Sarkozy and Hollande than to Chirac. Of course, this conclusion only applies to the end-of-the year addresses and would need to be confirmed for all the speeches.

All this shows that ChatGPT is a good imitator. Intertextual distance is therefore no longer able to identify texts generated by ChatGPT (as it once could be with older text generators) at least if users have taken care to submit to the generator a single homogeneous text to be "emulated"[6].

## 8   Conclusion

ChatGPT generates short texts, certainly because its prime vocation is a chatbot and not a public writer. When it comes to vocabulary and sentences, these generated texts display many singular characteristics, namely underuse of verbs (especially when conjugated in tenses other than the present), underuse of pronouns (especially third persons and indefinites), and underuse of adverbs and subordinating conjunctions. On the other hand, ChatGPT tends to overuse common nouns, adjectives, possessive determiners, and prepositions.

When looking at the sentence length and its distribution, the generator seems incapable of reproducing the diversity of sentence lengths, a characteristic of natural texts. Of course, these obvious limitations are not insurmountable and may be mitigated in future versions of the generator.

However, despite these limitations, when ChatGPT is placed in the best possible conditions (a single and short model, a known author, a date, a simple genre), it manages to reproduce the main formal characteristics of a natural text that has been given to it as a model. Moreover, the produced text makes sense to the reader who consults it. If this reader has any suspicions, there is currently no computerized tool to help confirm them. In fact, the techniques used until now to detect plagiarisms or texts generated by paper mills, no longer seem appropriate. Provided that users have taken care to submit homogeneous texts to ChatGPT, intertextual distance seems inoperative. Similarly, conventional plagiarism detection systems are likely to be ineffective, since they analyze n-grams (or word stacks), whereas GPT rearranges the vocabulary of the model in the generated text.

Further experiments will be needed to determine which features could be used to detect the generated texts. Moreover, experiments with other languages could confirm our findings. Finally, low-resource languages could represent a real challenge for LLMs due to their limited text corpora available.

---

[6] We view this term as preferable to "copied" or "imitated".